\documentclass[conference]{IEEEtran}
\IEEEoverridecommandlockouts
\usepackage{cite}
\usepackage{amsmath,amssymb,amsfonts}
\usepackage{algorithmic}
\usepackage{graphicx}
\usepackage{caption}
\usepackage{multirow}
\usepackage{amsmath}

\usepackage{breqn}

\usepackage{subcaption}
\usepackage{textcomp}
\usepackage{xcolor}
\def\BibTeX{{\rm B\kern-.05em{\sc i\kern-.025em b}\kern-.08em
    T\kern-.1667em\lower.7ex\hbox{E}\kern-.125emX}}
\begin{document}

\title{Social Impressions of the NAO Robot and its Impact on Physiology\\

}

\author{\IEEEauthorblockN{Ruchik Mishra}
\IEEEauthorblockA{\textit{Department of Electrical and Computer Engineering} \\
\textit{University of Louisville}\\
Louisville, USA \\
r0mish02@louisville.edu}
\and
\IEEEauthorblockN{Karla Conn Welch}
\IEEEauthorblockA{\textit{Department of Electrical and Computer Engineering} \\
\textit{University of Louisville}\\
Louisville, USA \\
karla.welch@louisville.edu}
}

\maketitle

\begin{abstract}
The social applications of robots possess intrinsic challenges with respect to social paradigms and heterogeneity of different groups. These challenges can be in the form of social acceptability, anthropomorphism, likeability, past experiences with robots etc. In this paper, we have considered a group of neurotypical adults to describe how different voices and motion types of the NAO robot can have effect on the perceived safety, anthropomorphism, likeability, animacy, and perceived intelligence of the robot. In addition, prior robot experience has also been taken into consideration to perform this analysis using a one-way Analysis of Variance (ANOVA). Further, we also demonstrate that these different modalities instigate different physiological responses in the person. This classification has been done using two different deep learning approaches, 1) Convolutional Neural Network (CNN), and 2) Gramian Angular Fields on the Blood Volume Pulse (BVP) data recorded. Both of these approaches achieve better than chance accuracy ($>$25\%) for a 4 class classification.


\end{abstract}


\section{Introduction}
Autism Spectrum Disorder (ASD), as an umbrella term, has been associated with challenges in social communication and interaction along with restrictive and repetitive behaviors according to the Diagnostic Statistical Manual \cite{american2013diagnostic}. The world ASD population is around 1.5 percent \cite{thabtah2020new}, with 1 out of 54 children being diagnosed in the United States alone \cite{alfuraydan2020use,baio2014prevalence}. More recently, this number has increased by 104\% \cite{saral2022autism}. Since, there is no cure for autism \cite{bolte2014autism, saral2022autism}, there exist numerous treatments/interventions, as presented by the authors in \cite{sandbank2020project,francis2005autism}, not all of which are scientifically proven to have positive results \cite{singer2015complementary,lilienfeld2014persistence,leaf2021evidence}. 

To mitigate the scientific shortcomings of some of these studies, Evidence Based Practices (EBP) have presented numerous interventions that have been chosen as a result of positive scientific evidence \cite{steinbrenner2020evidence}. Among them, Technology-Aided Instruction and Intervention (TAII) has been on the rise since a few decades now as is evident from the works of \cite{scassellati2012robots,dautenhahn2004towards, dautenhahn2003towards,werryz2001can}, some of which have also been used to form the guidelines for EBP like \cite{fridenson2017emotiplay,so2018using}.  
\begin{figure}[h!]
    \centering
    \includegraphics[scale=0.45]{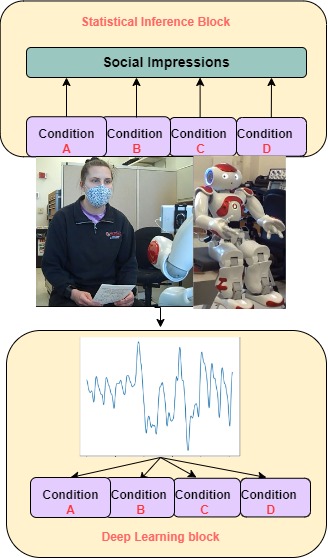}
    \caption{Social impressions of the NAO robot during HRI and its impact on physiology. Statistical Inference Block evaluates the importance of different modalities of the robot with respect to perceived features (pf). The Deep Learning block presents a time series classification approach to differentiate between the difference in physiology during different conditions A-D.}
    \label{fig:teaser_image}
\end{figure}
Apart from following the EBP, it is also essential to take into account the acceptability of the robot in a therapeutic scenario that involves a human-robot interaction (HRI). Due to the heterogeneity of the ASD population, these responses can vary based on the individual \cite{van2020adherence}. In addition to considering the social acceptability of the robot, factors such as repeated exposure to a robot \cite{robins2004effects}, features of robots to be considered in therapy \cite{lee2012robot}, anthropomorphic appearance and intonation \cite{van2018effects} etc. have also been studied. The reason for these studies can be attributed to the fact that these interventions are directed towards bringing about positive effects in the individual with ASD.

\begin{figure*}
   \begin{subfigure}[b]{0.45\textwidth}
         \centering
    \includegraphics[width=\textwidth]{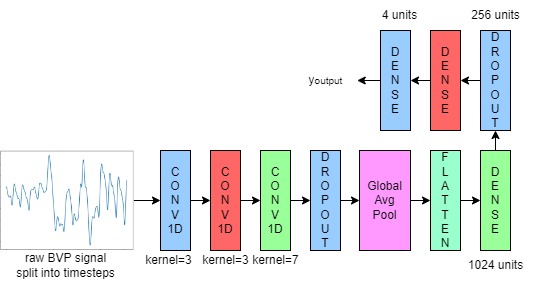}
         \caption{CNN architecture on the raw signal}
         \label{fig:raw_signal_cnn}
     \end{subfigure}
     \hfill
     \begin{subfigure}[b]{0.55\textwidth}
         \centering
         \includegraphics[width=\textwidth]{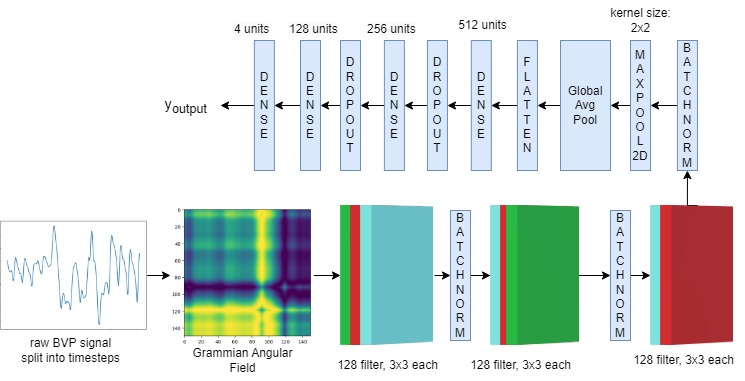}
         \caption{CNN architecture with Gramian Angular Field}
         \label{fig:GAF_CNN}
     \end{subfigure}
     \hfill
\caption{Deep learning models used for the two approaches. Figure \ref{fig:raw_signal_cnn} shows the model input of the raw signals. Figure \ref{fig:GAF_CNN} shows the use of Gramian Angular Fields with the CNN model.}
         \label{fig:model_Architectures}
\end{figure*}

Another aspect that can be associated with social acceptability of the robot during therapy is the uncanny valley, as has 
been studied by the authors in \cite{ueyama2015bayesian}.  Since the human-like traits of a robot can stir discomfort in people surrounding 
it; this phenomena, introduced in \cite{mori1970bukimi}, has been explored using different modalities of HRI \cite{mishra2022uncanny}. Since the effect of the uncanny valley has been shown to be more profound on the ASD population than on neurotypicals \cite{destephe2014uncanny}, it becomes imperative to validate individual communication modality in a robotic intervention to be analyzed on the neurotypical population first. Since the uncanny valley is associated with the features of the robots in terms of the appearance and its anthropomorphic appeal, it can have an 
effect on the affective states of the individual \cite{van2018effects}.

In this work, we have presented a one-to-one HRI as shown in Figure \ref{fig:teaser_image} with neurotypical adults. We start with the use of the humanoid NAO robot for a neurotypical population with four different conditions (A-D) (see Section \ref{impressions_of_nao} for more details). The participants' prior experience with robots has also been considered in this study. We have analysed the response of the participants on five pf of the robot namely, perceived safety, anthropomorphism, likeability, animacy, and perceived intelligence. We have analysed if the different conditions (A-D) have any effect on the users' perception of the five perceived robot features. In addition, we have compared the effects of varying amounts of prior experience with robots among the participants on pf under the four conditions (A-D). Further, the effect of these conditions on the physiological signals have been differentiated using deep learning algorithms.  

This paper has been arranged in the following way: Section \ref{Methodology} describes the methodology used in this paper followed by the results in Section \ref{results}. Further, Section \ref{limitations} outlines the future questions we would
like to address in our study followed by the conclusion in Section \ref{conclusion}.

\section{Methodology}\label{Methodology}

\subsection{Data acquisition}
We have considered 30 neurotypical adults for this study. The mean age of all the participants was 21.4 years with a standard deviation of 3.36 years. Out of the 30 adults, 43\% were male participants and the rest were female participants. During the sessions, the participants' audio, video and physiology was recorded. The physiological signals were recorded using the Empatica E4 wrist band. For the purpose of this study, just the BVP data was used for our deep learning based classification approaches. Before conducting this study, it was approved by the University's Institute Review Board (IRB). In addition, consent was taken from all the participants before the activity. There was no compensation involved for the participants involved in this research.

\subsection{Impressions of the NAO robot}\label{impressions_of_nao}
The participants were given a set of pre-defined questions to use to interact with the NAO robot. The modalities of the NAO robot were tested under four different scenarios: 
\begin{itemize}
    \item  Default NAO voice with smooth motions (Condition A), 
    \item Amazon Polly Justin voice [26] with smooth motions (Condition B), 
    \item Default NAO voice with jerky motions 
(Condition C), 
\item Amazon Polly Justin voice with jerky 
motions (Condition D). 
\end{itemize}
The participants' reactions to their impressions of the NAO robot were then assessed through the 
Godspeed and Robotic Social Attributes Scale (RoSAS) \cite{carpinella2017robotic} questionnaires across five categories: 1) perceived safety (PS), 2) anthropomorphism (AP), 3) animacy (AM), 4) likeability (LK) and, 5) perceived intelligence (PI), similar to the authors in \cite{bartneck2009measurement}. The scores for these categories 
were recorded after each of the robot conditions (A-D). Since the participants had varying backgrounds in terms of prior experience with robots 
in their personal or professional lives, this difference in experience with robots has been 
accounted for while assessing perceived safety, anthropomorphism, 
animacy, likeability, and perceived intelligence. The experience of a participant with robots was recorded on 
a scale of 0-3, where 0 stands for no familiarity whereas 3 stands for intermediate level of familiarity (e.g., completed projects with robots). All of 
these statistical inferences were made using the one-way Analysis of Variance (ANOVA) \cite{liu2015comparing,tamhane1977multiple} as has been discussed in later sections.

\subsection{Effect on Physiology}
Since the BVP data is recorded during
HRI, we use it to examine whether these physiological responses are differentiable using our deep learning algorithm. The BVP data is collected using a windowing approach. Since we use two different approaches for classification from the literature, the problem formulation for both of them have been explained.

\subsubsection{Problem formulation}
The first approach comprises of just CNNs for time series classification of the BVP signal. Given the univariate BVP signal data, $\mathcal{X} = \left\{x_{1}, x_{2} \dots , x_{n}\right\}$, split the time series into windows as has been shown in equation 
\begin{subequations}\label{eq:window_signal}
\begin{equation}
  f_{\text{split}} (\mathcal{X}) = \mathcal{Y}
\end{equation}    
\begin{equation}
 \Rightarrow g_{i}(x) =   \left\{x_{i*j}, \dots , x_{i*j + \text{p}} \right\} \rightarrow \mathcal{Y}_{i}
\end{equation}
\end{subequations}
where $i$ is the number of training data point made from splitting the time series each into $p$ steps, $j$ denotes the stride length we used, and $\mathcal{Y}$ denotes the labels w.r.t. the conditions A-D. This basic formulation is used to find the multivariate function $g(x)$ using our proposed Deep Learning approaches.

\subsubsection{Deep Learning for classification}
The first approach used for our classification problem is based on CNNs on the raw time series signal which has been split into smaller windows. The network architecture has been shown in Figure \ref{fig:model_Architectures}. For the second architecture, instead of using raw signals as inputs, we convert the split signals from equation \ref{eq:window_signal} into images based on Gramian Angular Fields \cite{wang2015imaging,xu2020human}. In this work, we use the Gramian Angular Field Difference (GADF) as defined in\cite{wang2015imaging} as:
\begin{dmath}
    \text{GADF}_{i} = \left(\sqrt{ I -\tilde{X}^{2}_{[i*j:i*j + \text{p}]}}\right)^{T} . \tilde{X}_{[i*j:i*j + \text{p}]} - (\tilde{X}_{[i*j:i*j + \text{p}]})^{T}. \sqrt{ I -\tilde{X}^{2}_{[i*j:i*j + \text{p}]}}
\end{dmath}
where $\tilde{X}$ are the polar coordinates of the time-series BVP signal $\mathcal{X}$, and $I$ is the unit vector \cite{wang2015imaging}. The definition of the polar coordinate $\tilde{X}$ can be expressed as in \cite{wang2015imaging}:
    \begin{dmath}
        \left\{\tilde{x}_{i}| \tilde{x}_{i} \in \tilde{X}_{i}, -1 \leq \tilde{x}_{i} \leq 1, (\tilde{x}^{i}_{-1}, \tilde{x}^{i}_{0}) = \left( \frac{(x_{i}-\text{max}(\mathcal{X})+x_{i}-\text{min}(\mathcal{X}))}{\text{max}(\mathcal{X}) - \text{min} (\mathcal{X})} , \\ \frac{x_{i} -  \text{min}(\mathcal{X}) }{  \mathcal{X} - \text{min}(\mathcal{X})     }                    \right)            \right\} ,  \textbf{where } x_{i} \in \mathcal{X}
    \end{dmath}
In the above equation, the values of the BVP signal $\mathcal{X}$ can be scaled between $[-1,1]$ as has been mentioned in the expression of $\tilde{x}^{i}_{-1}$ or between $[0,1]$ as has been mentioned by the expression of $\tilde{x}^{i}_{0}$.

\section{Results and Discussions}\label{results}

\begin{figure*}[h!]
     \centering
     \begin{subfigure}[b]{0.30\textwidth}
         \centering
    \includegraphics[width=\textwidth]{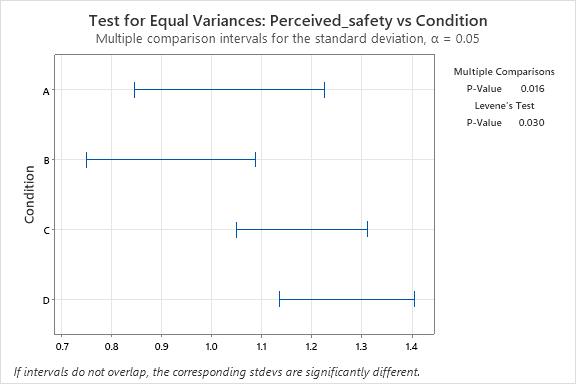}
         \caption{Perceived safety}
         \label{fig:PS_v}
     \end{subfigure}
     \hfill
     \begin{subfigure}[b]{0.30\textwidth}
         \centering
    \includegraphics[width=\textwidth]{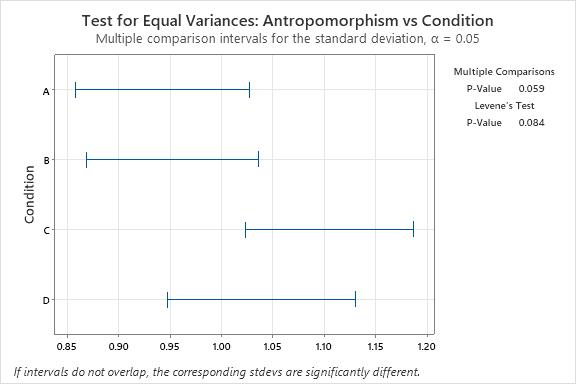}
         \caption{Anthropomorphism}
         \label{fig:AP_v}
     \end{subfigure}
     \hfill
     \begin{subfigure}[b]{0.30\textwidth}
         \centering
    \includegraphics[width=\textwidth]{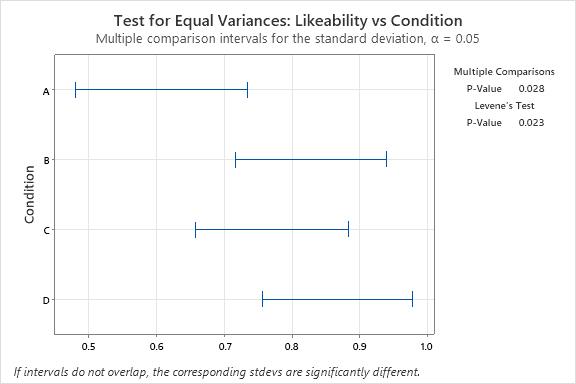}
         \caption{Likeability}
         \label{fig:LK_v}
     \end{subfigure}
       \begin{subfigure}[b]{0.30\textwidth}
         \centering
    \includegraphics[width=\textwidth]{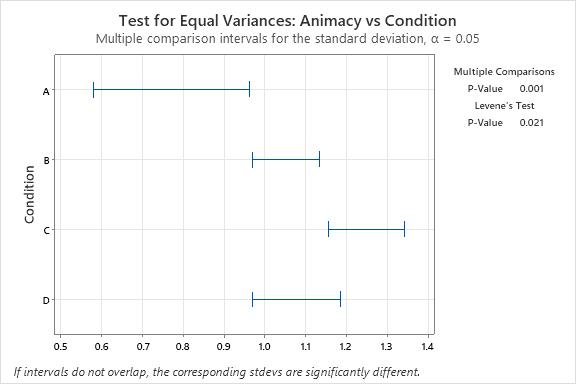}
         \caption{Animacy}
         \label{fig:PI_v}
     \end{subfigure}
      \begin{subfigure}[b]{0.30\textwidth}
         \centering
    \includegraphics[width=\textwidth]{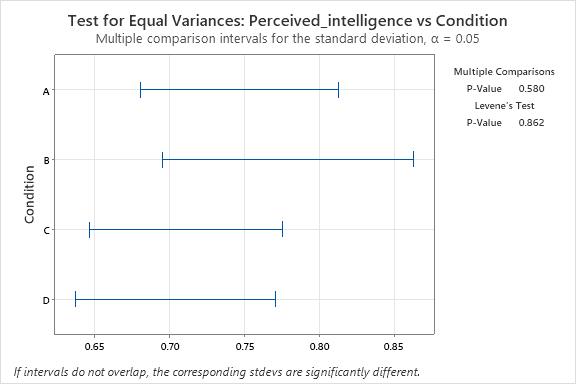}
         \caption{Perceived Intelligence}
         \label{fig:PI_v}
     \end{subfigure}
      \begin{subfigure}[b]{0.35\textwidth}
         \centering
    \includegraphics[width=\textwidth]{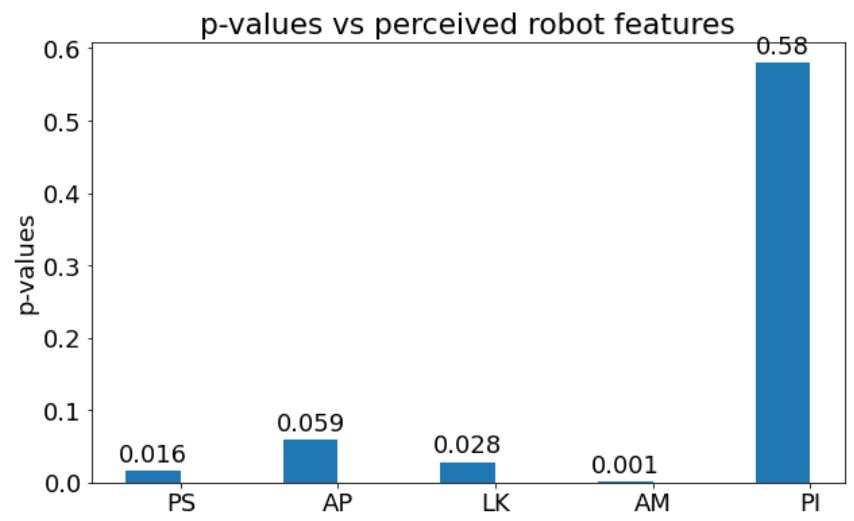}
         \caption{$p$-values}
         \label{fig:PI_v}
     \end{subfigure}
        \caption{Test for equal variance in the populations for conditions A, B, C, and D each for perceived safety, anthropomorphism, likeability, animacy, and perceived intelligence.}
        \label{fig:test_variance}
\end{figure*}

\subsection{Effects of Conditions (A-D) on user experience}
\subsubsection{Test for equal variances}
The first step is to check the null hypothesis ($H_{o}$) in equation \ref{eq:null} for each of the features: PS, AP, AM, LK and, PI.

\begin{figure*}[h!]
     \centering
     \begin{subfigure}[b]{0.30\textwidth}
         \centering
    \includegraphics[width=\textwidth]{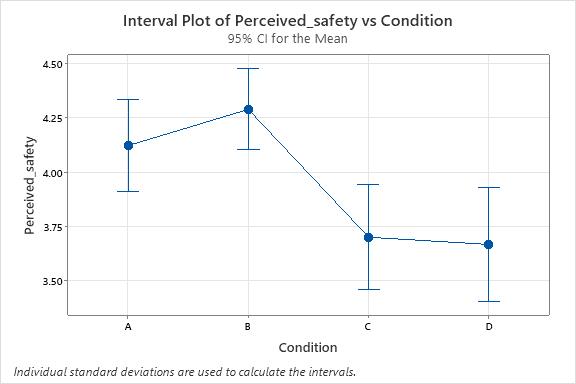}
         \caption{Perceived safety}
         \label{fig:PS_AW}
     \end{subfigure}
     \hfill
     \begin{subfigure}[b]{0.30\textwidth}
         \centering
    \includegraphics[width=\textwidth]{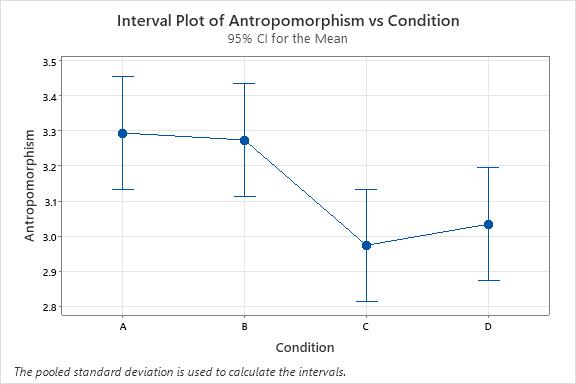}
         \caption{Anthropomorphism}
         \label{fig:AP_AW}
     \end{subfigure}
     \hfill
     \begin{subfigure}[b]{0.30\textwidth}
         \centering
    \includegraphics[width=\textwidth]{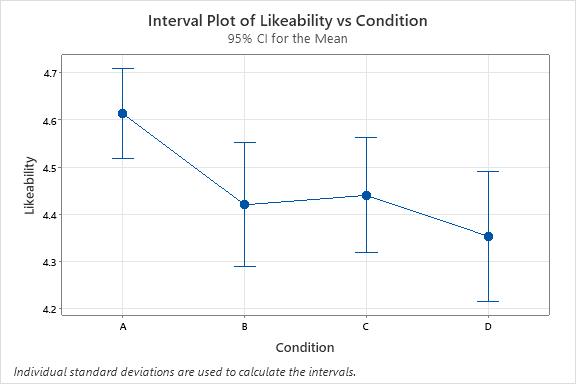}
         \caption{Likeability}
         \label{fig:LK_AW}
     \end{subfigure}
       \begin{subfigure}[b]{0.30\textwidth}
         \centering
    \includegraphics[width=\textwidth]{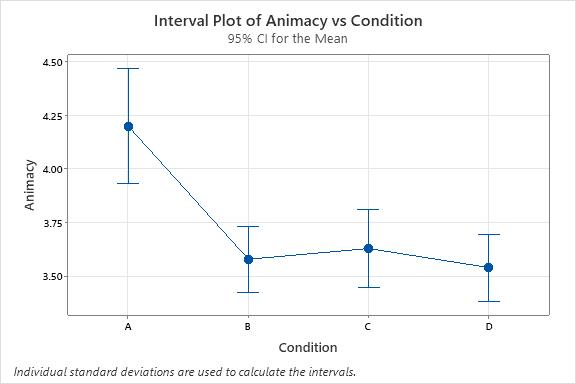}
         \caption{Animacy}
         \label{fig:AN_AW}
     \end{subfigure}
      \begin{subfigure}[b]{0.30\textwidth}
         \centering
    \includegraphics[width=\textwidth]{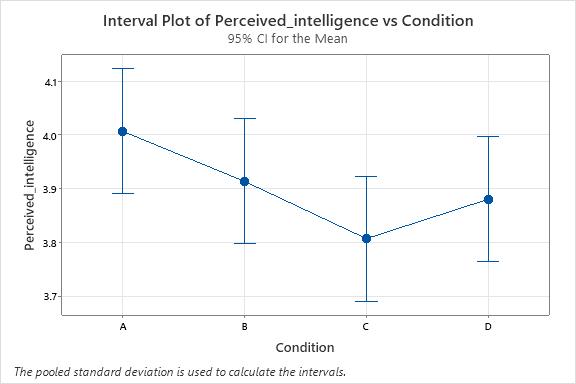}
         \caption{Perceived Intelligence}
         \label{fig:PI_AW}
     \end{subfigure}
      \begin{subfigure}[b]{0.38\textwidth}
         \centering
    \includegraphics[width=\textwidth]{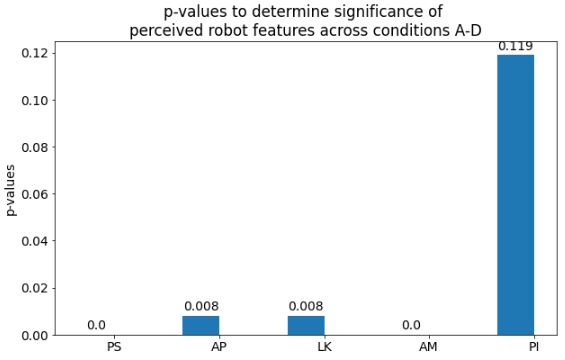}
         \caption{$p$-values}
         \label{fig:p_AW}
     \end{subfigure}
        \caption{One-way ANOVA for finding significance of conditions A-D for different perceived robot features.}
        \label{fig:ANOVA_whole}
\end{figure*}

\begin{subequations}\label{eq:null}
\begin{equation}
    H_{o}: \sigma_{A}^{2} = \sigma_{B}^{2} = \sigma_{C}^{2} = \sigma_{D}^{2}
\end{equation}
\begin{equation}
    H_{1}: \exists \sigma_{i} \neq \sigma_{j} \textrm{ for } i\neq j \textrm{ where } i,j = \left\{A,B,C,D\right\}
\end{equation}
\end{subequations}

\begin{figure*}[h!]
     \centering
     \begin{subfigure}[b]{0.40\textwidth}
         \centering
    \includegraphics[width=\textwidth]{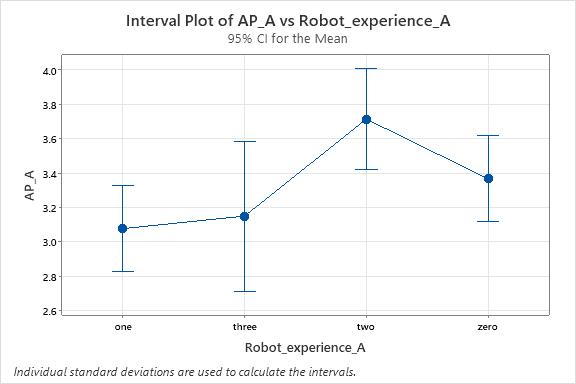}
         \caption{Anthropomorphism under condition A is significant}
         \label{fig:AP_re}
     \end{subfigure}
     \hfill
     \begin{subfigure}[b]{0.40\textwidth}
         \centering
    \includegraphics[width=\textwidth]{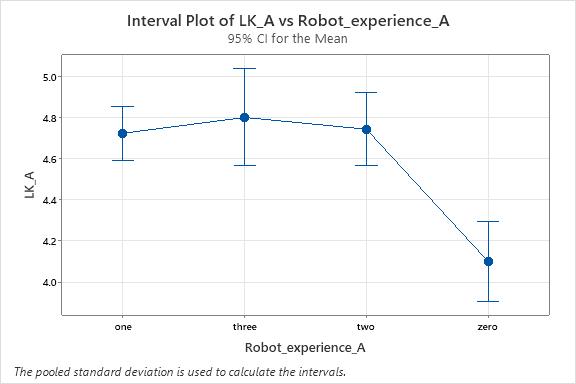}
         \caption{Likeability under condition A is significant}
         \label{fig:LKA_re}
     \end{subfigure}
     \hfill
     \begin{subfigure}[b]{0.40\textwidth}
         \centering
    \includegraphics[width=\textwidth]{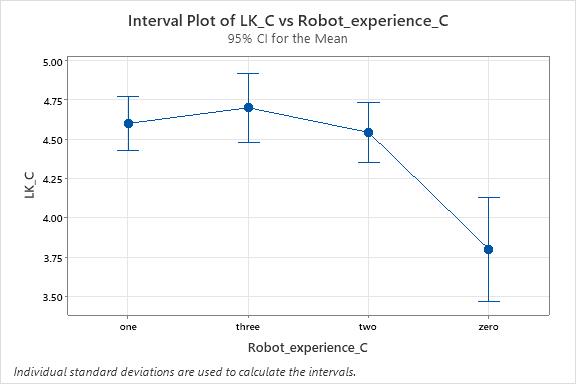}
         \caption{Likeability under condition C is significant}
         \label{fig:LKC_re}
     \end{subfigure}
       \begin{subfigure}[b]{0.40\textwidth}
         \centering
    \includegraphics[width=\textwidth]{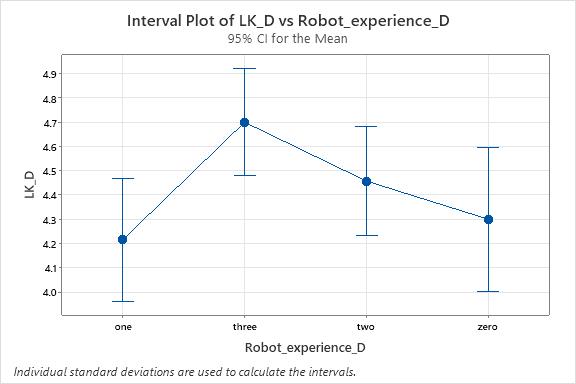}
         \caption{Likeability under condition D is significant}
         \label{fig:LKD_re}
     \end{subfigure}
        \caption{Effect of experience with robotics on different perceived features under different conditions.}
        \label{fig:RE}
\end{figure*}
\begin{table*}[htbp]
  \caption{Robot experience}
  \label{tab:robot_exp}
  \centering
  \begin{tabular}{|c|c|c|c|c|c|}
    \hline
    \multicolumn{2}{|c|}{\multirow{3}{*}{\textbf{Perceived features}}} & \multicolumn{4}{|c|}{\textbf{$p-$values}} \\
    \cline{3-6}
    \multicolumn{2}{|c|}{} & \textbf{Condition A} & \textbf{Condition B} & \textbf{Condition C} & \textbf{Condition D} \\
    \hline
    \multirow{2}{*}{Perceived Safety} & Test of variance & 0.912 &0.643  & 0.733 & 0.136\\
    \cline{2-6}
                           & One-way ANOVA & 0.592 &0.237 & 0.083 & 0.376  \\
    \hline
    \multirow{2}{*}{Anthrpomorphism} & Test of variance & \textbf{0.042} & 0.468 & 0.640 & 0.055 \\
    \cline{2-6}
                           & One-way ANOVA & \textbf{0.013} & 0.411 & 0.548 & 0.288 \\
 \hline
   \multirow{2}{*}{Likeability} & Test of variance & 0.586 & 0.112 & \textbf{0.042} & \textbf{0.005} \\
    \cline{2-6}
                           & One-way ANOVA & \textbf{0.000} & 0.206 & \textbf{0.000} & \textbf{0.025} \\
 \hline
   \multirow{2}{*}{Animacy} & Test of variance & 0.073 & 0.071 & 0.901 & 0.295 \\
    \cline{2-6}
                           & One-way ANOVA & 0.911 & 0.359 & 0.268 & 0.433 \\                         
    \hline
  \end{tabular}
\end{table*}
From the $p$-values, obtained using Bonferroni 95\% confidence intervals for standard deviations, we can reject the null hypothesis for PS, LK, and AM ($p$-value$\leq 0.05$) which means that the variances for these groups are different \cite{hommel1988stagewise,gordi2004simple}. On the other hand, we accept the null hypothesis for anthropomorphism and 
perceived intelligence ($p$-value$> 0.05$),  which means that for each of these features, we can consider the variances of conditions A-D to be equal.

\subsubsection{ANOVA for each perceived feature}
For this we consider the data of all the 30 participants included in this study irrespective of their previous exposure to robots. We use Welch's one-way ANOVA \cite{liu2015comparing,tamhane1977multiple} for perceived safety, likeability, and animacy since we do not assume equal variances as was shown from the results in Figure \ref{fig:test_variance}. We compare the difference between conditions A-D by making a hypothesis similar to that in equation \ref{eq:null}, but this time with the means of the values:
\begin{subequations}\label{eq:feature_anova}
\begin{equation}
    H_{o}: \mu_{A} = \mu_{B} = \mu_{C} = \mu_{D}
\end{equation}
\begin{equation}
    H_{1}:  \exists \mu_{i} \neq \mu_{j} \textrm{ for } i\neq j \textrm{ where } i,j = \left\{A,B,C,D\right\}
\end{equation}
\end{subequations}

\begin{figure*}
   \begin{subfigure}[b]{0.32\textwidth}
         \centering
    \includegraphics[width=\textwidth]{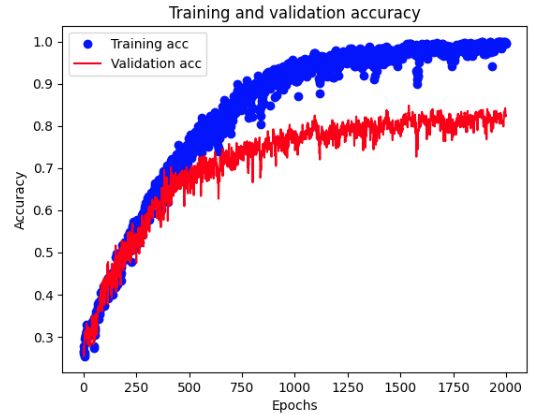}
         \caption{Train and validation accuracy. $p=512$, $j = 128$. Test acc = $0.8179$}
         \label{fig:option2_raw}
     \end{subfigure}
     \hfill
     \begin{subfigure}[b]{0.32\textwidth}
         \centering
         \includegraphics[width=\textwidth]{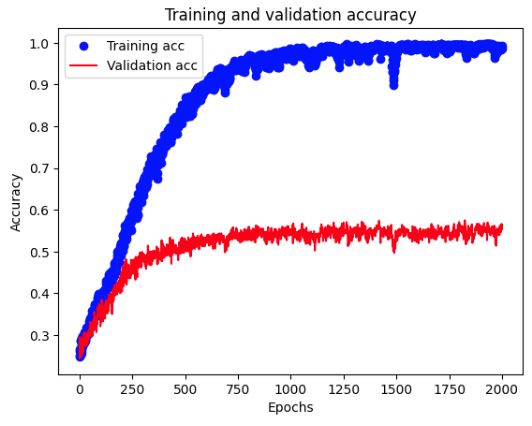}
         \caption{Train and validation accuracy. $p=256$, $j = 128$. Test acc = $0.5831$}
         \label{fig:option3_raw}
     \end{subfigure}
      \begin{subfigure}[b]{0.32\textwidth}
         \centering
         \includegraphics[width=\textwidth]{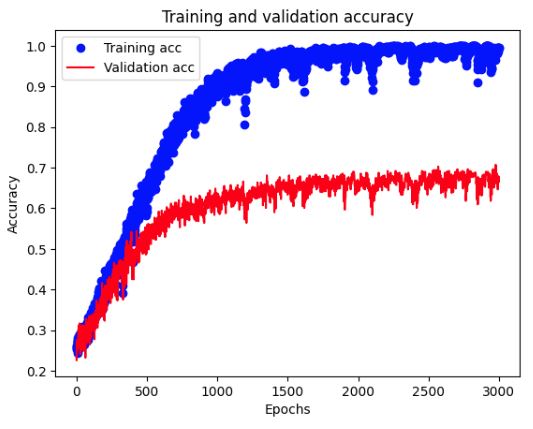}
         \caption{Train and validation accuracy. $p=512$, $j = 200$. Test acc = $0.6767$}
         \label{fig:option5_raw}
     \end{subfigure}
     \hfill
\caption{Effect of using different sliding window size and stride length on the accuracy of classification using the raw singal.}
         \label{fig:accuracy_raw_signal}
\end{figure*}

\begin{figure*}
   \begin{subfigure}[b]{0.32\textwidth}
         \centering
    \includegraphics[width=\textwidth]{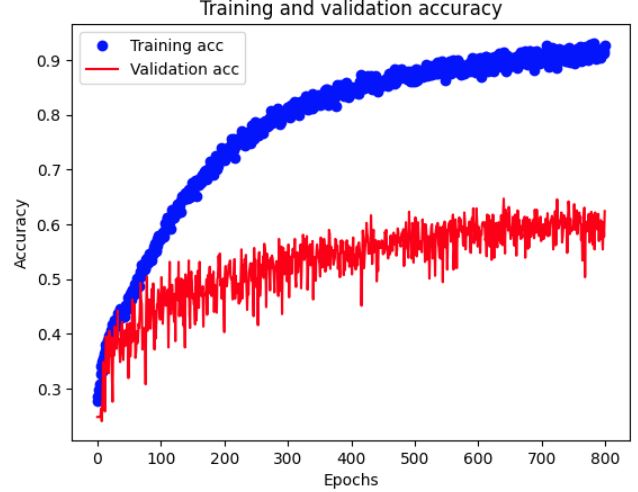}
         \caption{Train and validation accuracy. $p=200$, $j = 50$. Test acc = $0.5988$}
         \label{fig:option2_raw}
     \end{subfigure}
     \hfill
     \begin{subfigure}[b]{0.32\textwidth}
         \centering
         \includegraphics[width=\textwidth]{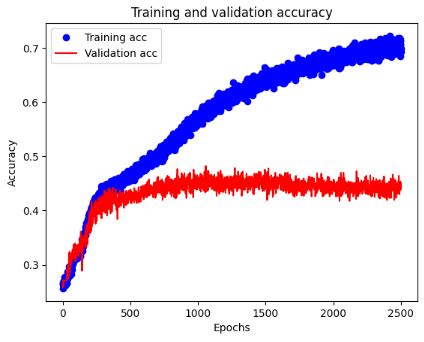}
         \caption{Train and validation accuracy. $p=150$, $j = 30$. Test acc = $0.4599$}
         \label{fig:option3_raw}
     \end{subfigure}
      \begin{subfigure}[b]{0.32\textwidth}
         \centering
         \includegraphics[width=\textwidth]{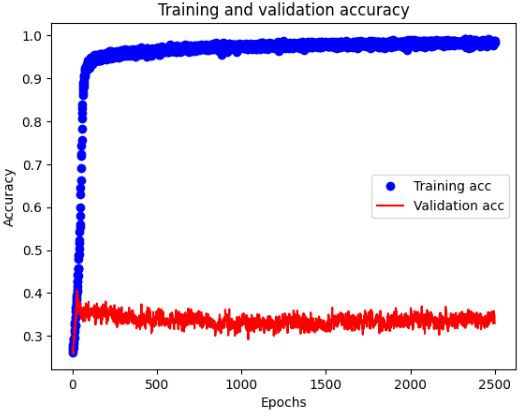}
         \caption{Train and validation accuracy. $p=512$, $j = 128$. Test acc = $0.3505$}
         \label{fig:option5_raw}
     \end{subfigure}
     \hfill
\caption{Effect of using different sliding window size and stride length on the accuracy of classification using Gramian Angular Field.}
         \label{fig:acc_gaf}
\end{figure*}

The main effects plot for each of the perceived features has been shown in Figure \ref{fig:PS_AW}-\ref{fig:PI_AW}, and the $p$-values for them have been shown in Figure \ref{fig:p_AW}. From the $p-$values it can be seen that for perceived safety, anthropomorphism, likeability and animacy, we can reject the null hypothesis $(H_{o})$ since the $p$-value $\leq0.05$, whereas we accept $H_{o}$ for perceived intelligence ($p$-value $>0.05$). This means that under different conditions A-D, a user has different levels of perception of PS, AP, LK, and AM; but the perceived intelligence of the robot remains unaffected. From the main effects plot in Figures \ref{fig:PS_AW}-\ref{fig:AN_AW}, we can also conclude that:
\begin{itemize}
    \item \underline{Perceived safety}: B$>$A$>$C$>$D. This means that the modality of the Amazon Polly Justin voice of the NAO robot with smooth motions was perceived to be the safest among users. 
    
    \item \underline{Anthropomorphism}: A$>$B$>$D$>$C. The users perceived the NAO robot's default voice with smooth motions to be most anthropomorphic (human-like).
    \item \underline{Likeability}: A$>$C$>$B$>$D. The participants liked the NAO's default voice with its smooth motions most, similar to the case of anthropomorphism.
    \item \underline{Animacy}:  A$>$B$>$C$>$D. Condition A still has the most preference for animacy as compared to the other conditions.
\end{itemize}

\subsection{Effects of Experience with robots}
In this section, we compare the effects of varying robot experiences for each of the significantly perceived features ($p-$value $\leq$ 0.05) evaluated from the previous section (i.e perceived safety, anthropomorphism, likeability, and animacy). We have considered 4 levels of robot experiences:
\begin{itemize}
    \item \textbf{0: } No experience/never heard of robots before today
    \item \textbf{1: } Fundamental awareness (basic knowledge/awareness of robots)
    \item \textbf{2: } Novice (limited direct experience with robots)
    \item \textbf{3: } Intermediate (completed practical applications with robots)
\end{itemize}
We use a similar hypothesis as in equation \ref{eq:feature_anova} for finding the effect of robot experience on the perceived features:
\begin{subequations}\label{eq:robot_Experience}
\begin{equation}
    H_{o}: \mu_{\textbf{0}} = \mu_{\textbf{1}} = \mu_{\textbf{2}} = \mu_{\textbf{3}}
\end{equation}
\begin{equation}
    H_{1}: \exists \mu_{i} \neq \mu_{j} \textrm{ for } i\neq j \textrm{ where } i,j = \left\{0,1,2,3\right\}
\end{equation}
\end{subequations}
The details of test of variances, $p-$values for all combinations of conditions and features have been shown in Table \ref{tab:robot_exp}. It can be seen that prior experience with robots does not have any effect on the participants under any of the conditions (A-D) as far as perceived safety of robots during the interaction is concerned. The same holds for animacy too. On the other hand, under some conditions (A-D), anthropomorphism and likeability of the robot depend on prior experience with robotics has been highlighted in Table \ref{tab:robot_exp}. The main effects plot of these significant values have been shown in Figure \ref{fig:RE}. Based on these graphs, we conclude that for condition A, novice participants (level-\textbf{`2'}) found the robot to be more anthropomorphic than the other participants. Similarly, participants with experience level-\textbf{`3'} found the robot more likeable for conditions A, C, and D.

\subsection{Effects on Physiological response}
From equation \ref{eq:window_signal}, we can see that the sliding window approach has $j$ strides in between two consecutive windows. The approach has been adopted from the authors in \cite{reiss2019deep} where an eight-second sliding window is used with a two-second stride length. We analyzed the effect of different stride length on the train, test, and validation accuracy of both of our models. All the code for these deep learning models was run on Google Colab's premium version. The learning rate used for each of these simulations was 0.001 with Adam optimization and accuracy as the metric. The framework used for this work was Tensorflow.

\subsubsection{CNN model with raw BVP signal}
The effective length, as defined by the equation in \ref{eq:effective_len}, has been used as the reference for inspecting different lengths of the sliding window and strides.
\begin{equation}\label{eq:effective_len}
    \text{Effective length} = \frac{j}{p}
\end{equation}
As can be seen from Figure \ref{fig:accuracy_raw_signal}, the effective length of 0.25 (Figure \ref{fig:option2_raw}) fetched the maximum test accuracy (0.8179) followed by the effective length of 0.39 (Figure \ref{fig:option3_raw}) with a test accuracy of 0.6767 and effective length of 0.5 with a test accuracy of 0.5831. In addition, the validation accuracies are also in agreement with the test accuracies for each of the effective length values. Hence, as the effective length increases, the validation and test accuracies of the model decreases.  

\subsubsection{CNN model with Gramian Angular Field}
The training and test accuracy for this case has been shown in Figure \ref{fig:acc_gaf}. As can be seen from the low test accuracies of using the Gramian Angular Field, the effective length did not have any influence on the test accuracies. This can be attributed to the trend stationarity of the BVP signal \cite{akar2013spectral}. 
\subsubsection{Proof of trend stationarity of BVP signal}
We have used the Augmented Dickey-Fuller (ADF) test \cite{baum2018kpss} and the Kwiatkowski-Phillips-Schmidt-Shin (KPSS) Test \cite{mushtaq2011augmented} for testing the trend stationarity of the BVP signal.
\begin{table}[htbp]
  \caption{Statistical test for stationarity}
  \label{tab:robot_exp}
  \centering
  \begin{tabular}{ |p{2cm}|p{1cm}|p{1cm}|p{1cm}|p{1cm}| }
\hline
\multicolumn{5}{|c|}{$p$-values w.r.t. 95\% confidence interval} \\
\hline
\textbf{Tests} & \textbf{A} & \textbf{B} & \textbf{C} & \textbf{D} \\
\hline
\hline
ADF & 0.1 & 0.1 & 0.1 & 0.1\\
\hline
KPSS & 0.0 & 0.0 & 0.0 & 0.0\\
\hline
\end{tabular}
\end{table}
Equation \ref{eq:statistics_time_series} shows the hypothesis as suggested in \cite{web:AnalyticsVidya}.
\begin{subequations}\label{eq:statistics_time_series}
\begin{equation}
    H_{o}: \text{Time series has a unit root}
\end{equation}
\begin{equation}
    H_{1}:  \text{Time series has no unit root}
\end{equation}
\end{subequations}
For ADF test, we can reject the Null hypothesis ($H_{o}$) as $p$-value $=0.1>0.05$, whereas in case of KPSS test, we accept $H_{o}$ ($p$-value $=0<0.05$). Hence, the BVP signal is trend-stationary \cite{web:AnalyticsVidya} which is consistent with the observations by authors in \cite{akar2013spectral}.

\section{Limitations and Future Work}\label{limitations}
For the user acceptance part of this paper, we would like to further test the impressions of children with ASD to the robot's modalities for 
comparing the perceived features of the NAO robot. In addition, we would like to expand the capability of the NAO robot to be able to understand and respond to candid conversations rather than pre-defined questions which is aligned with the aims in \cite{mishra2022towards}. For the classification problem described, we would like to include the use of autoencoders for denoising the BVP signal \cite{huamin2020reconstruction,zheng2022denoising} towards making an end-to-end deep learning pipeline for classification.


\section{Conclusion}\label{conclusion}
In this paper, we present a study that involves the use of the humanoid NAO robot for a neurotypical population with four different conditions (A-D) of varying voice types and motions during HRI (see Section \ref{impressions_of_nao} for more details). The participants' prior experience with robots has also been considered in this study. We have analysed the response of the participants on five perceived features (pf) of the robot namely, perceived safety, anthropomorphism, likeability, animacy, and perceived intelligence. We have analysed if the different conditions (A-D) have any effect on the users' perception of the five perceived robot features. In addition, we have compared the effects of varying amounts of prior experience with robots among the participants on pf under the four conditions (A-D). In the end, we demonstrated the effect of these conditions on the physiology (here BVP) of the participants. Based on the performance of our deep learning approach, we were able to classify the physiological responses of the participants under different conditions with more than chance ($>25\%$) accuracy. Between the two approaches used for classification, using the raw signals with a CNN model (test acc: 0.8179 \%) worked better than using GAF (test accuracy: 0.59\%) attributing to the trend stationarity of the BVP signal.


\section*{Ethical Impact Statement}
As described in Section \ref{Methodology}, prior IRB approval was taken before conducting this study. Informed consent was taken from all the participants. In addition, they had an option to opt out of the study at any stage.

The data collected during this study had no gender or racial bias. We had a close to equal male to female ratio (Male= 43\%, Female= 57\%). However, efforts were not made to ensure cross cultural data collection. Additional research would be needed to address the social impressions of the NAO robot across cultures.

For the deep learning model used, the generalization is performed across the subject data collected currently. Although, only BVP data was used as a physiological marker for differentiating between the conditions A-D. This distinction might not be generalizable across different physiological signals like Electrodermal Activity (EDA) or temperature.




\section*{Acknowledgment}\label{acknowledgements}
The authors wish to acknowledge undergraduate researchers for their assistance with collecting the data. We also want to thank the adult subjects and the staff at the university-affiliated autism center.


\bibliographystyle{plain}
\bibliography{references.bib}

\end{document}